\documentclass[pdflatex,sn-nature]{sn-jnl}

\usepackage{graphicx}%
\usepackage{multirow}%
\usepackage{amsmath,amssymb,amsfonts}%
\usepackage{amsthm}%
\usepackage{mathrsfs}%
\usepackage[title]{appendix}%
\usepackage{xcolor}%
\usepackage{textcomp}%
\usepackage{manyfoot}%
\usepackage{booktabs}%
\usepackage{algorithm}%
\usepackage{algorithmicx}%
\usepackage{algpseudocode}%
\usepackage{listings}%

\unnumbered 

\begin{document}

\title[A unified framework for geometry-independent operator learning in cardiac electrophysiology]{A unified framework for geometry-independent operator learning in cardiac electrophysiology}


\author[1]{\fnm{Bei} \sur{Zhou}}
\author[1]{\fnm{Cesare} \sur{Corrado}}
\author[1]{\fnm{Shuang} \sur{Qian}}
\author[1,2]{\fnm{Maximilian} \sur{Balmus}}
\author[1]{\fnm{Angela W. C.} \sur{Lee}}
\author[1]{\fnm{Cristobal} \sur{Rodero}}
\author[3]{\fnm{Caroline} \sur{Roney}}
\author[4]{\fnm{Marco J.W.} \sur{Götte}}
\author[4]{\fnm{Luuk H.G.A.} \sur{Hopman}}
\author[6]{\fnm{Gernot} \sur{Plank}}
\author[5]{\fnm{Mengyun} \sur{Qiao}}
\author[1, 2]{\fnm{Steven} \sur{Niederer}}

\affil[1]{\orgname{Imperial College London},  \orgaddress{\city{London}, \country{United Kingdom}}}
\affil[2]{\orgname{Alan Turing Institute}, \orgaddress{\city{London}, \country{United Kingdom}}}
\affil[3]{\orgname{Queen Mary University of London},  \orgaddress{\city{London}, \country{United Kingdom}}}
\affil[4]{\orgname{Amsterdam University Medical Center}, \orgaddress{\city{Amsterdam}, \country{The Netherlands}}}
\affil[5]{\orgname{University College London},  \orgaddress{\city{London}, \country{United Kingdom}}}
\affil[6]{\orgname{Medical University of Graz},  \orgaddress{\city{Graz}, \country{Austria}}}


\abstract{
Learning neural operators on heterogeneous and irregular geometries remains a fundamental challenge, as existing approaches typically rely on structured discretisations or explicit mappings to a shared reference domain. We propose a unified framework for geometry-independent operator learning that reformulates the learning problem in an intrinsic coordinate space defined on the underlying manifold. By expressing both inputs and outputs in this shared coordinate domain, the framework decouples operator learning from mesh discretisation and geometric variability, while preserving meaningful spatial organisation and enabling faithful reconstruction on the original geometry.

We demonstrate the framework on cardiac electrophysiology, a particularly challenging setting due to extreme anatomical variability across heart geometries. Leveraging a GPU-accelerated simulation pipeline, we generate large-scale datasets of high-fidelity electrophysiology simulations across diverse patient-specific anatomies and train customised neural operators to predict full-field local activation time maps. The proposed approach outperforms established neural operators on both atrial and ventricular geometries. Beyond cardiac electrophysiology, we further show that the same representation enables operator learning in cardiac biomechanics, a distinct problem involving volumetric deformation, highlighting the generality of the proposed framework. Together, these results establish intrinsic coordinate representations as a principled and extensible pathway for neural operator learning on complex physical systems characterised by heterogeneous geometry.
}

\maketitle

\section{Introduction}

Learning solution operators for partial differential equations on complex geometries remains a fundamental challenge across many areas of computational science. Neural operators offer a promising paradigm by learning mappings between functional spaces, enabling solution fields to be predicted for unseen parameters, discretisations, or geometries without retraining \cite{kovachki2023neural}. Architectures such as DeepONet \cite{don} and the Fourier Neural Operator (FNO) \cite{fno} have shown strong performance on parametric PDE problems, particularly on structured or weakly deformed domains. However, extending these approaches to heterogeneous, irregular geometries remains difficult, as most neural operator formulations implicitly rely on consistent spatial discretisations or globally aligned coordinate systems \cite{geofno, gino}.

A central obstacle is that real-world physical systems are often defined on manifolds with substantial variability in shape, topology, and mesh representation. When learning is performed directly in the space of mesh discretisations, geometric differences become entangled with the underlying physical solution, hindering generalisation across geometries and limiting the applicability of neural operators to patient-specific or instance-specific domains. Existing approaches address this challenge by deforming irregular geometries to reference domains or by encoding geometry as auxiliary input \cite{yin2024scalable, zhao2025diffeomorphism}. While effective for relatively simple shapes, such strategies can become unstable or inaccurate for domains with complex topology, multiple boundaries, or substantial variability across instances.

Here, we propose a general framework for geometry-independent neural operator learning that reformulates the learning problem in a coordinate domain intrinsic to the underlying manifold, rather than in the space of mesh discretisations. By expressing model inputs and outputs in a shared intrinsic parameterisation, spatial solution fields can be represented consistently across heterogeneous geometries despite substantial variability in shape, topology, and discretisation. Learning is performed entirely in this intrinsic coordinate domain, while predictions are subsequently mapped back to the original physical geometry for faithful reconstruction and interpretation. Importantly, the framework itself only assumes the existence of a smooth intrinsic coordinate chart on the manifold, making it broadly applicable beyond any specific physical system.

Cardiac electrophysiology serves as a challenging and clinically relevant testbed for this framework, due to extreme anatomical variability, complex topology, and the need for high spatial fidelity. Computational models of cardiac electrophysiology are widely used to study electrical wave propagation, investigate arrhythmia mechanisms, and inform patient-specific therapeutic interventions \cite{niederer2019computational, trayanova2024computational, prakosa2018personalized}. A key output of such models is the local activation time (LAT) map, which characterises the spatiotemporal evolution of electrical activation across the myocardium \cite{williams2018local, jaffery2024review, karoui2021cardiac}. Generating LAT maps typically requires large ensembles of high-fidelity finite element simulations on anatomically detailed, patient-specific meshes, making conventional workflows computationally expensive and difficult to scale \cite{niederer2011verification, opencarp}.

In this work, we adopt the Universal Atrial Coordinates (UAC) \cite{uac} and Universal Ventricular Coordinates (UVC) \cite{uvc} as concrete instantiations of intrinsic coordinate representations for atrial and ventricular geometries, respectively. These coordinate systems preserve physiologically meaningful spatial organisation while providing a unified domain across diverse anatomies, enabling neural operators to learn spatially coherent solution mappings without sensitivity to irregular mesh connectivity or topological inconsistencies. To support operator learning at scale, we further introduce an efficient GPU-accelerated electrophysiology simulation pipeline that enables the generation of large ensembles of biophysically detailed LAT maps across diverse patient-specific geometries.

By integrating intrinsic coordinate representations with high-performance simulation and neural operator learning, the proposed framework addresses both the geometric heterogeneity and the computational cost that have limited prior approaches. While cardiac electrophysiology serves as a challenging and clinically relevant demonstration, the framework is broadly applicable to other physical systems in which intrinsic coordinates are available, illustrating a general pathway toward geometry-independent operator learning on complex manifolds.

\section{Results}
\subsection{Large-scale electrophysiology datasets in intrinsic coordinate space}

We first assess whether an intrinsic coordinate representation yields a consistent learning domain across heterogeneous geometries, using LAT map prediction in cardiac electrophysiology as a representative operator learning task.

We generated a comprehensive set of biophysically detailed EP simulations using a custom GPU-accelerated FEM solver applied to two independent cohorts of AF patients (Cohort~A: $n=100$; Cohort~B: $n=47$). This process constructed two large-scale datasets, Dataset A and Dataset B, respectively (see Methods). To ensure robustness across diverse anatomical and physiological conditions, for each of the $147$ patient-specific left atriums (LAs), we simulated LAT maps from seven distinct pacing sites across the LA. At each site, we sampled $300$ combinations of the key conduction parameters, the longitudinal conductivity ($\sigma_l$) and the anisotropy ratio ($\sigma_l/\sigma_t$), using Latin Hypercube Sampling (LHS) \cite{mckay2000comparison} to cover a wide range of clinically relevant tissue properties. This yields a total of 308,700 simulated LAT maps, establishing one of the largest computational datasets of its kind for training neural operators.

\begin{figure}[ht]
  \centering
  \includegraphics[width=1.0\linewidth]{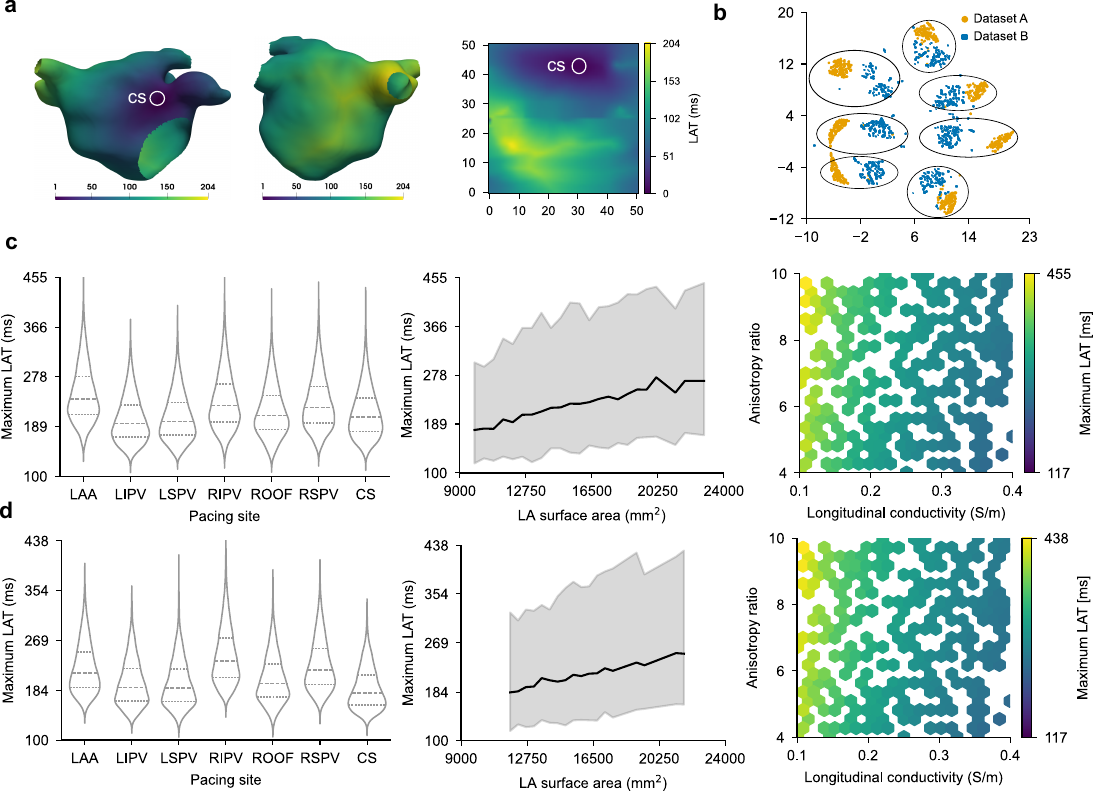}
    \caption{\textbf{Construction and characterisation of large-scale electrophysiology datasets generated from biophysically detailed simulations.}
    \textbf{a}, Example local activation time (LAT) map from a 3D left atrium (LA) electrophysiology simulation with coronary sinus pacing, together with its projection onto the standardised Universal Atrial Coordinate (UAC) domain.
    \textbf{b}, UMAP embedding of all UAC-projected LAT maps from Datasets~A and~B, showing seven clusters per cohort corresponding to pacing sites and systematic displacement between corresponding clusters, indicating a pronounced domain shift between the two datasets.
    \textbf{c}, Distribution of maximum LAT across pacing locations in Dataset~A, demonstrating linear dependence on LA surface area and sensitivity to conductivity.
    \textbf{d}, Corresponding analysis for Dataset~B, showing the same physiological trends despite cohort-level differences. }

  \label{fig:datasets}
\end{figure}

All simulated LAT maps were subsequently projected onto the standardised UAC domain to obtain a consistent, geometry-independent representation. Fig.~\ref{fig:datasets}a illustrates an LAT map on the 3D atrial surface together with its corresponding 2D UAC projection, with the coronary sinus pacing site producing full atrial activation within 204 ms.

To characterise variability within this intrinsic representation, we performed dimensionality reduction on the full set of UAC-projected LAT maps. High-dimensional LAT fields were first denoised using principal component analysis (PCA), and the resulting principal components were subsequently embedded using Uniform Manifold Approximation and Projection (UMAP)~\citep{healy2024uniform}. The 2D embedding (Fig.~\ref{fig:datasets}b) reveals a clear separation between the two datasets. Within each cohort, seven well-defined clusters emerge, each corresponding to one of the pacing sites. Clusters associated with the same pacing location in Datasets~A and~B do not overlap but instead form displaced pairs in the latent space. This consistent displacement provides visual evidence of a domain shift between the cohorts, arising not from physiological differences, but from systematic variations in anatomical distributions, acquisition protocols, or preprocessing pipelines. This domain shift has a measurable impact on cross-domain generalisation performance. The methodology used to assess this effect is detailed in Methods. 

We next quantified the primary determinants of total activation time across all simulations. Three factors exert the strongest influence: pacing site, LA surface area, and conduction velocities. The maximum LAT varies across pacing locations in Dataset~A (Fig.~\ref{fig:datasets}c) and Dataset~B (Fig.~\ref{fig:datasets}d). Across both datasets, maximum LAT increases approximately linearly with LA surface area and decreases with increasing conductivity values. These consistent trends confirm that surface area and conductivity jointly modulate global activation timing, motivating their inclusion as key anatomical features in the subsequent analyses.

\subsection{Sensitivity to intrinsic representation and input features}

\begin{figure}[ht]
  \centering
  \includegraphics[width=1.0\linewidth]{figs/results.pdf}
  \caption{\textbf{Comprehensive evaluation of model performance for LAT map prediction.}
    \textbf{a}, Our model performance from feature sensitivity tests on Dataset A, demonstrating the contribution of each input feature.
    \textbf{b}, Evaluation of the impact of two regularisation terms, namely total variance (TV) and Laplacian, on LAT map prediction on Dataset A.
    \textbf{c}, Performance comparison between our proposed model and established deep learning architectures, including DeepONet, Fourier Neural Operator (FNO), Wavelet Neural Operator (WNO), U-Net, and ResNet on Dataset A.
    \textbf{d}, Spatial distribution of the averaged LAT prediction error (in ms) across five test cases in a representative validation fold. White circles highlight regions with the highest prediction errors, and adjacent numbers show the average error within those specific regions.
    \textbf{e}, Performance comparison between Single Pacing Location Models and a Unified Model across seven pacing sites. 
    \textbf{f}, Cross-domain performance comparison on two distinct datasets. The notation $X \rightarrow Y$ indicates that the model was trained on source dataset $X$ and evaluated on target dataset $Y$. }
  \label{fig:results}
\end{figure}

We quantified the contribution of individual anatomical input modalities by removing spatial coordinates, fibre orientation, and surface area features, retraining the model under each configuration. Model performance was evaluated using mean absolute error (MAE), which captures numerical accuracy in milliseconds, and the structural similarity index measure (SSIM) \cite{wang2004image}, which assesses spatial fidelity of the predicted activation maps. All results are reported using five-fold cross-validation. Unless otherwise stated, experiments were conducted on Dataset A, with cross-domain evaluations involving both Datasets A and B.

As shown in Fig. \ref{fig:results}a, incorporating all anatomical features yields accurate and spatially coherent predictions of local activation time. Removing the surface area feature leads to a substantial degradation in performance, highlighting the importance of global anatomical scaling for correctly organising activation wavefronts. Similarly, excluding spatial coordinates significantly reduces both numerical accuracy and structural similarity, demonstrating that explicit localisation in the UAC domain is essential for resolving conduction pathways.

In contrast, removing fibre orientation features unexpectedly improves predictive performance across both metrics. This observation indicates that, in the present formulation, fibre orientation does not provide an informative signal beyond that already captured by geometric coordinates and global anatomical features. Consequently, fibre orientation inputs were excluded from all subsequent experiments. This finding does not imply that fibre orientation is unimportant for cardiac electrophysiology, but rather that its current atlas-based representation, when projected into the intrinsic coordinate domain, does not contribute additional predictive value for this task.

\subsection{Effect of spatial regularisation on operator learning}

We investigated the impact of spatial regularisation on LAT map reconstruction by augmenting the $H^1$ loss with two commonly used smoothness terms: total variation (TV) and a second-order Laplacian penalty, applied individually or in combination. These regularisation terms are designed to suppress high-frequency artefacts while preserving physiologically meaningful spatial structure.

Model performance was evaluated across a range of regularisation strengths ($\lambda \in \{5 \times 10^{-2}, 3 \times 10^{-2}, 1 \times 10^{-2}, 5 \times 10^{-3}, 1 \times 10^{-3}\}$) using five-fold cross-validation. For each configuration, performance metrics were averaged across folds and regularisation weights to isolate the effect of the regularisation formulation (Fig.~\ref{fig:results}b).

Among all tested configurations, combining the $H^1$ loss with both TV and Laplacian regularisation consistently yielded the best performance, achieving the lowest mean MAE and highest SSIM. This result indicates that jointly enforcing first- and second-order smoothness constraints enhances structural fidelity without sacrificing numerical accuracy. In contrast, applying either TV or Laplacian regularisation alone resulted in smaller and less consistent improvements, suggesting that their combined effect is necessary to adequately capture the spatial organisation of activation patterns.

\subsection{Comparison with existing neural operator architectures}

We benchmarked our approach against a range of neural operators and convolutional architectures commonly used for modelling cardiac activation from multimodal inputs. To ensure a fair comparison, models were evaluated across approximately capacity-matched configurations spanning representative hyperparameter ranges (Fig.~\ref{fig:results}c). Full architectural details and exhaustive results are provided in Supplementary Section~2 and Supplementary Table~1.

Among operator learning methods, the Fourier Neural Operator (FNO) and Wavelet Neural Operator (WNO) exhibited limited empirical performance in this setting, indicating challenges in capturing heterogeneous and noisy electrophysiological patterns on complex anatomical domains. DeepONet, implemented using our ViT-based backbone for both branch and trunk networks, achieved substantially improved accuracy, reflecting its stronger capacity to model global functional relationships.

Convolutional baselines showed a similar trend. The U-Net architecture outperformed the simpler ResNet model, consistent with its multi-scale encoder–decoder structure and strong inductive bias for spatial localisation. Nevertheless, convolutional models remained limited in their ability to capture global pacing effects and long-range spatial dependencies.

Our model outperformed all baselines, achieving the highest numerical accuracy and structural fidelity across all evaluated metrics. This result demonstrates its superior ability to integrate heterogeneous anatomical and physiological inputs into spatially coherent and physiologically consistent LAT predictions.

To assess where residual errors concentrate spatially, we visualised prediction error distributions across the UAC domain (Fig.~\ref{fig:results}d). Across representative test cases, errors are predominantly localised to a narrow horizontal band corresponding to the UAC seam, while remaining low and diffuse elsewhere, indicating high fidelity across most anatomical regions.

Beyond accuracy, we evaluated computational efficiency to assess suitability for real-time applications. On a modern NVIDIA GPU, the model achieved an average inference time of approximately $0.12$~ms per sample. This represents orders of magnitude acceleration relative to conventional FEM electrophysiology solvers, enabling near real-time LAT prediction across diverse anatomies and pacing configurations.

\subsection{Generalisation across stimulation conditions}

Generalisation across diverse pacing conditions is critical for scalable and clinically useful prediction of atrial activation, as it determines whether a single model can match or exceed the performance of models trained for individual pacing sites. A model that performs well across pacing sites indicates that it has learned shared electrophysiological structure rather than site-specific patterns.

We thus assessed the predictive performance of two modelling paradigms for LAT map reconstruction: (i) Single Pacing Location Models, trained independently for each of seven pacing sites, and (ii) a Unified Model, trained jointly across all sites to learn a shared inductive bias over anatomical and physiological structure. Across all pacing conditions, the Unified Model yields consistent and statistically significant improvements over its independently trained counterparts. Specifically, it achieves a 16.5\% reduction in mean MAE from 6.3 ms to 5.2 ms (Fig.~\ref{fig:results}e, left), demonstrating enhanced predictive accuracy across heterogeneous inputs. The model also delivers a 1.67\% increase in average SSIM from 0.960 to 0.976 (Fig.~\ref{fig:results}e, right), reflecting superior preservation of spatial structure and clinically relevant activation patterns. 

\subsection{Generalisation under latent domain shifts} 
\label{sec:cross_domain}

Our initial experiments confirmed the strong performance of our model on a single dataset (Dataset A), demonstrating its ability to capture anatomical features relevant to the target task within a controlled domain. However, high performance within the same dataset does not guarantee translational reliability. Here, we distinguish between internal performance, which measures the model’s accuracy when trained and tested on the same dataset, and cross-domain performance, which measures generalisation to an independent dataset. Evaluating both is critical to determine whether the model has learned biologically meaningful features or relies on dataset-specific cues.

Cross-domain generalisation was evaluated on two levels. First, we measured model performance when trained and tested on the same dataset (internal performance: $A \rightarrow A$ and $B \rightarrow B$) compared with performance when trained on one dataset and tested on the other (cross-domain performance: $A \rightarrow B$ and $B \rightarrow A$). Second, we assessed whether including data from an additional domain ($(A + B) \rightarrow A$, $(A + B) \rightarrow B$, and $(A + B) \rightarrow (A + B)$) during training improves generalisation to the original dataset, using internal performance as a baseline. 

Each evaluation was performed on models trained with 18 combinations of hyperparameters to ensure statistical significance, with results highlighting consistent trends (Fig.~\ref{fig:results}f; full results are provided in Supplementary Tables~2 and~3). The strong internal performance confirms that the UAC projection effectively separates anatomical representation from patient-specific shape. In contrast, the pronounced drop in cross-domain performance provides clear evidence of latent bias. The model relies on image-specific cues that mislead predictions when applied to a new domain. This finding highlights a second-order challenge in translational modelling. Models are highly sensitive to systematic variations in data representation and annotation rather than true anatomical differences.

Integrating heterogeneous datasets during training mitigates these domain-specific biases. Training on the combined datasets $(A + B)$ encourages the model to focus on consistent anatomical features, ignoring spurious correlations tied to individual acquisition or annotation pipelines. Across both datasets, MAE decreased, and SSIM increased in comparison with the internal performance, demonstrating that exposure to independent cohorts acts as a form of regularisation and that geometry-independent representations alone are insufficient for translational robustness unless coupled with heterogeneous, multi-domain training. For completeness, we also evaluated the model trained and tested on the combined datasets $(A + B) \rightarrow (A + B)$, obtaining $(5.77 \pm 0.16)$ for MAE and $(0.9734 \pm 0.003)$ for SSIM.  

\subsection{Reconstruction on original geometries}

\begin{figure}[ht]
  \centering
  
  \includegraphics[width=1.0\linewidth]{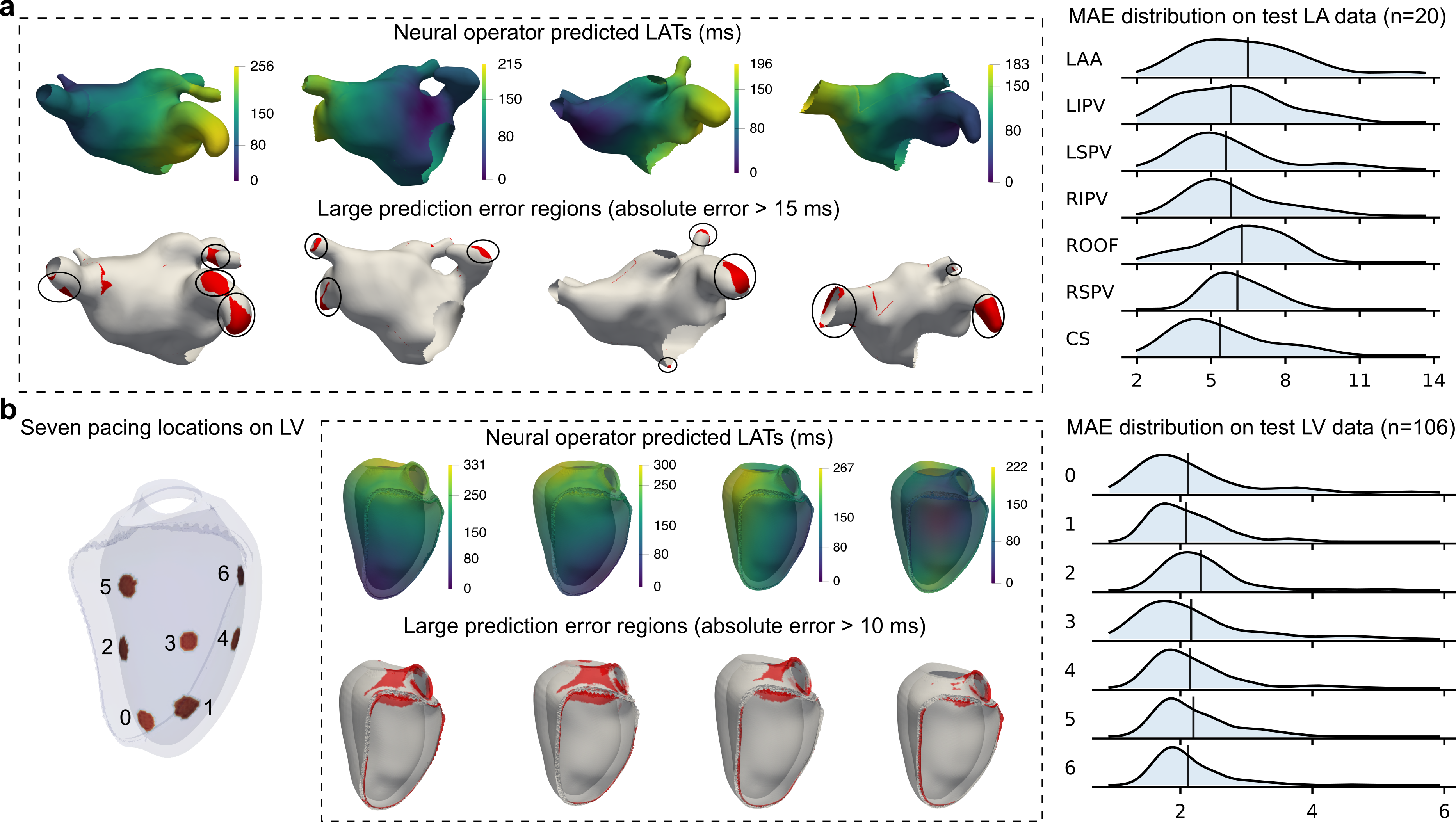}
  \caption{\textbf{Interpolation of neural operator predictions to patient-specific atrial and ventricular mesh geometries.}
\textbf{a}, Left: local activation time (LAT) predictions interpolated from the coordinate domain back onto representative left atrial (LA) surface meshes. Regions where the absolute prediction error exceeds 15~ms are highlighted. Right: distribution of mean absolute error across pacing locations evaluated on 100 LA test cases.
\textbf{b}, Corresponding results for left ventricular (LV) meshes, showing interpolated LAT predictions, regions of large absolute error, and error distributions across pacing locations.}
\label{fig:onmesh}
\end{figure}

To assess anatomical prediction fidelity after interpolation, we evaluated model performance on both LA surface meshes and left ventricular (LV) volume meshes (Fig.~\ref{fig:onmesh}). Predictions produced in the intrinsic anatomical coordinate domain were interpolated back onto the original patient-specific mesh geometry, enabling anatomically faithful reconstruction in both settings.

Model performance was evaluated using a previously proposed benchmark task in which LAT is predicted from one of seven pacing locations alone \cite{yin2024scalable}. Our experiments were conducted on Cohort~A using 100 LA geometries, of which 20 were held out for testing. In addition, 1,006 LV meshes were obtained from a publicly available dataset \cite{lvdata}, with 106 meshes reserved for evaluation.  Model inputs and outputs were represented on a $100\times100$ grid for LAs and a $50\times50\times50$ volumetric grid for LVs.

On test data, the model achieves mean absolute errors of 5–6 ms for LA geometries, relative to maximum activation times of up to 285 ms. For LV geometries, errors are consistently lower, typically ranging between 2 and 4 ms for maximum activation times of up to 350 ms. These ventricular error levels improve upon those reported in prior geometry-independent operator learning approaches, where mean absolute errors of approximately 5–6 ms were observed for LV meshes \cite{yin2024scalable}. The comparison is based on matched dataset scale, task definition, and activation time range, rather than a direct reimplementation, which is not feasible because the prior work relies on a simplified LV representation that is not compatible with the extracted LV meshes considered here.

In both atrial and ventricular settings, errors remain consistent across pacing locations, indicating stable performance independent of stimulation site. Spatially, elevated errors in the LA are primarily localised to the atrial appendage and pulmonary vein openings, regions characterised by complex geometry and boundary effects. For the LV, larger errors are observed near the basal openings, corresponding to truncation at the valve plane, and along the interface with the right ventricle, reflecting geometric discontinuities introduced during extraction from bi-ventricular meshes.

\subsection{Application to kinematic deformation modelling}

To showcase that the proposed approach is not specific to cardiac electrophysiology, we apply it to a numerically distinct problem: kinematic deformation of non-linear elastic solids under load. This task involves volumetric deformation governed by material stiffness and applied pressure, and therefore represents a fundamentally different class of physical system and boundary conditions.

We generate a dataset of three-dimensional geometries by varying two independent geometric parameters controlling global curvature and wall thickness. For each geometry, kinematic deformation simulations are performed across different combinations of material stiffness and internal pressure under fixed boundary constraints. As in the cardiac electrophysiology setting, a problem-specific intrinsic coordinate system is constructed to provide a consistent parameterisation across heterogeneous meshes, enabling all displacement fields to be represented in a shared coordinate domain. 

The neural operator takes four scalar inputs describing geometry and loading conditions and predicts the full-field displacement magnitude. Performance is evaluated using ten-fold cross-validation. The average displacement magnitude in the dataset is approximately $2\,\mathrm{mm}$, with a maximum of about $7\,\mathrm{mm}$, while the model achieves a mean absolute error of $0.35 \pm 0.28\,\mathrm{mm}$ across test folds. Relative to the characteristic displacement magnitudes in the dataset, this error level indicates that the learned operator captures the dominant deformation behaviour across heterogeneous geometries.

\section{Discussion}
\label{sec:discussion}

\subsection{Geometry-independent neural operator learning}

A key insight arising from this work is that geometry independence in operator learning can be most effectively addressed at the level of representation rather than architecture. Within this framework, the role of the intrinsic coordinate system is not to simplify the underlying physics, but to provide a canonical representation in which spatially coherent learning becomes feasible. By embedding geometric structure directly into the learning representation, neural operators can exploit both local spatial relationships and global contextual information without being confounded by irregular mesh connectivity or patient-specific discretisation artefacts. Importantly, the framework itself is agnostic to the specific physical system, requiring only the availability of a smooth intrinsic coordinate chart that is consistent across instances. Potential applications extend to include fluid–structure interaction, geophysical modelling on planetary surfaces, and biological transport processes on irregular anatomical domains.

Cardiac electrophysiology provides a particularly stringent and clinically relevant testbed for this methodology, due to the pronounced variability in anatomy, topology, and boundary conditions across patients. By leveraging universal anatomical coordinate systems, we instantiate the proposed framework in this setting and demonstrate that neural operators can capture the complex, nonlinear dynamics of electrical conduction across highly heterogeneous cardiac geometries. The resulting prediction accuracy, with mean absolute errors on the order of 5 ms, lies within the typical measurement uncertainty and physiological variability reported for clinical electroanatomic mapping systems \cite{gaeta2021high} (typically 5--10 ms \cite{narayan2024advanced}). This indicates that the learned operators are not only numerically accurate but also operate within a clinically acceptable margin.

A further premise underlying geometry-independent operator learning is the need for large, diverse, and high-quality training data. To support this, we developed a pipeline featuring a GPU-accelerated finite element electrophysiology solver that enables efficient generation of high-fidelity simulations across a wide range of patient-specific left atrial anatomies and electrophysiological conditions. This data generation capability addresses a longstanding computational bottleneck in cardiac modelling and provides a scalable foundation for learning generalisable structure within the proposed framework.

Beyond enabling the present study, the resulting dataset establishes a resource for future data-driven investigations. Its scale, diversity, and internal consistency allow predictive models to generalise across anatomies and simulation protocols, a capability that has previously been constrained by data scarcity. For comparison, G-FuNK \cite{loeffler2024graph} was trained on 25 left atrial geometries, while DIMON \cite{yin2024scalable} leveraged 1,006 ventricular geometries simulated at multiple pacing sites, yielding 7,042 simulation instances. In contrast, our simulations preserve multiple pacing locations while also varying tissue conductivity, allowing neural operators to capture electrophysiological responses across diverse stimulation patterns and heterogeneous, patient-specific anatomies. Overall, the dataset comprises 308,700 simulations, offering a substantially richer foundation for training and benchmarking state-of-the-art neural operators.

\subsection{Inductive biases for learning on heterogeneous geometries}

The comparative evaluation reveals that strong performance in geometry-independent operator learning depends less on the choice of a specific neural architecture than on the inductive biases it encodes. In particular, accurate reconstruction of spatial fields on heterogeneous domains requires a balance between global contextual reasoning and precise local spatial representation. Models that emphasise only one of these aspects consistently underperform in this setting.

Our approach benefits from this balance by combining global context modelling with explicit spatial structure, enabling the network to capture long-range dependencies while preserving fine-scale localisation. This is particularly important for activation time prediction, where global wavefront propagation and sharp local gradients coexist within the same spatial field. Architectures that capture global dependencies without maintaining sufficient spatial fidelity struggle to resolve rapid local transitions, while those that prioritise local structure alone fail to model long-range interaction driven by wavefront propagation.

These trends are consistent across the evaluated baselines. Operator learning methods designed to model smooth global mappings perform well in terms of structural similarity but exhibit reduced pointwise accuracy in regions with steep spatial gradients. Conversely, convolutional architectures with strong local inductive biases generate visually coherent predictions yet remain limited in their ability to account for global pacing effects without substantial increases in depth or receptive field. Multi-resolution approaches offer partial mitigation but remain sensitive to alignment errors and domain irregularities in this context.

Together, these results indicate that geometry-independent operator learning on irregular anatomical domains places specific demands on model design. Architectures must simultaneously integrate global information and preserve spatial locality in order to achieve both numerical accuracy and structural fidelity. This requirement is largely independent of the specific application and is expected to generalise to other problems involving partial differential equations on complex geometries.

\subsection{Toward domain-invariant generalisation}

Generalisation across heterogeneous input conditions is a central challenge for data-driven modelling on complex anatomical domains. Prior studies have primarily evaluated performance under variations in pacing site \cite{yin2024scalable}. In contrast, our results show that a single model can generalise across both diverse pacing locations and simulation parameters. Models trained jointly across all pacing sites consistently outperform those trained for individual locations, indicating that the learned representations capture electrophysiological structure that is not tied to a specific stimulation configuration.

Beyond pacing variability, our cross-domain experiments reveal a more fundamental limitation. Performance degradation when models are trained on a single dataset and evaluated on an independent cohort is not primarily attributable to the choice of network architecture, but rather reflects latent heterogeneity in imaging, annotation, and preprocessing protocols. By explicitly combining datasets acquired under distinct conditions, the model is encouraged to discard protocol-specific shortcuts and instead focus on anatomical features that are consistent across domains. This strategy not only improves performance on external cohorts but also yields modest gains on internal test sets.

Together, these observations suggest a general strategy for achieving domain-invariant learning in medical imaging and simulation-based modelling. Universal coordinate representations provide a geometry-invariant foundation, while training on heterogeneous data acts as an implicit regularisation in feature space, promoting representations that reflect underlying biophysical structure rather than dataset-specific artefacts. More broadly, these results indicate that robust translational performance is more likely to arise from integrating diverse, multi centre data than from optimising models within a single acquisition domain.

\subsection{Limitations and future directions}

One limitation revealed by our ablation studies concerns the role of fibre orientation features. Removing fibre orientation may initially appear counterintuitive, yet it consistently improved predictive performance. Two factors are likely to contribute to this outcome. First, the current fibre orientation representation may introduce noise or uncertainty that interferes with learning the underlying electrophysiological dynamics. Second, the model appears to rely more effectively on the spatial and geometric priors provided by the anatomical coordinates and surface area features, which may already encode sufficient structural information. In this setting, explicit fibre orientation inputs may therefore be redundant or even detrimental in their present form.

Beyond feature representation, limitations also arise from the anatomical coordinate parameterisation itself. For the left atrium, the UAC system maps a closed, topologically complex three-dimensional surface onto a flat two-dimensional domain, which necessarily introduces a cut line that appears as a seam in the coordinate representation. As a result, points that are physically adjacent on the original anatomy become separated across this seam, creating an artificial spatial discontinuity for the learning model. Consistent with this construction, our results reveal a narrow band of elevated prediction error aligned with the UAC seam, indicating difficulty in interpolating smoothly across this boundary.

Additional localised errors are observed in regions of pronounced anatomical and electrophysiological complexity, particularly around the pulmonary vein junctions and the left atrial appendage. These regions exhibit interpatient variability that cannot be fully normalised by the UAC projection, leading to residual geometric heterogeneity in the coordinate domain. Together, these factors contribute to small but systematic clusters of elevated error in physiologically critical regions. While the proposed framework captures global activation patterns and structure over long spatial scales, resolving fine-scale anatomical variation remains challenging.

By contrast, ventricular geometries parameterised using the UVC do not exhibit a comparable global seam-induced discontinuity. Instead, ventricular error behaviour is governed primarily by boundary effects arising from geometric truncation rather than by coordinate-induced artefacts. Importantly, these truncations preserve the overall ventricular topology and continuity of the domain. This contrast highlights that limitations associated with intrinsic coordinate representations depend on underlying anatomical topology rather than being intrinsic to the operator learning framework itself, and that coordinate-based representations may therefore behave differently across cardiac chambers.

Finally, the framework was trained exclusively on high-fidelity simulated datasets. Although these simulations span a wide range of anatomies and electrophysiological parameters, validation against clinical electroanatomic mapping data will be essential to assess robustness under real-world conditions. In addition, the current implementation focuses primarily on atrial and ventricular geometries in isolation. Extending the approach to whole-heart or fully biventricular modelling will require adapting the anatomical parameterisation and scaling the data generation process. Future directions include integrating hybrid simulation–clinical datasets, refining anatomical representations in regions of high complexity, and exploring adaptive spatial resolution strategies to improve local accuracy while preserving computational efficiency and physical fidelity.

\section{Methods}

\subsection{Geometry-independent operator learning framework}

The proposed framework separates operator learning from geometric representation by introducing an intrinsic coordinate domain that serves as the sole space in which learning is performed. Both input fields and target outputs are represented on this intrinsic domain at a fixed resolution, which allows standard neural operator architectures defined on regular grids to be used directly. The original physical mesh is not used during learning itself and is used only to project quantities that depend on geometry and physical parameters into the intrinsic domain.

The framework consists of three stages. First, a problem-specific intrinsic coordinate map is used to project geometric coordinates, boundary information, and physical parameters from a mesh domain to a structured coordinate grid. Second, high-fidelity simulations are performed on the original mesh, and the resulting output fields are projected to the same coordinate grid to form paired training data, which are then used to train a neural operator in the coordinate domain.
Third, predictions produced on the coordinate grid are interpolated back to the original mesh to enable anatomically faithful visualisation and downstream analysis. 

\begin{figure}[htbp]
  \centering
  \includegraphics[width=\textwidth]{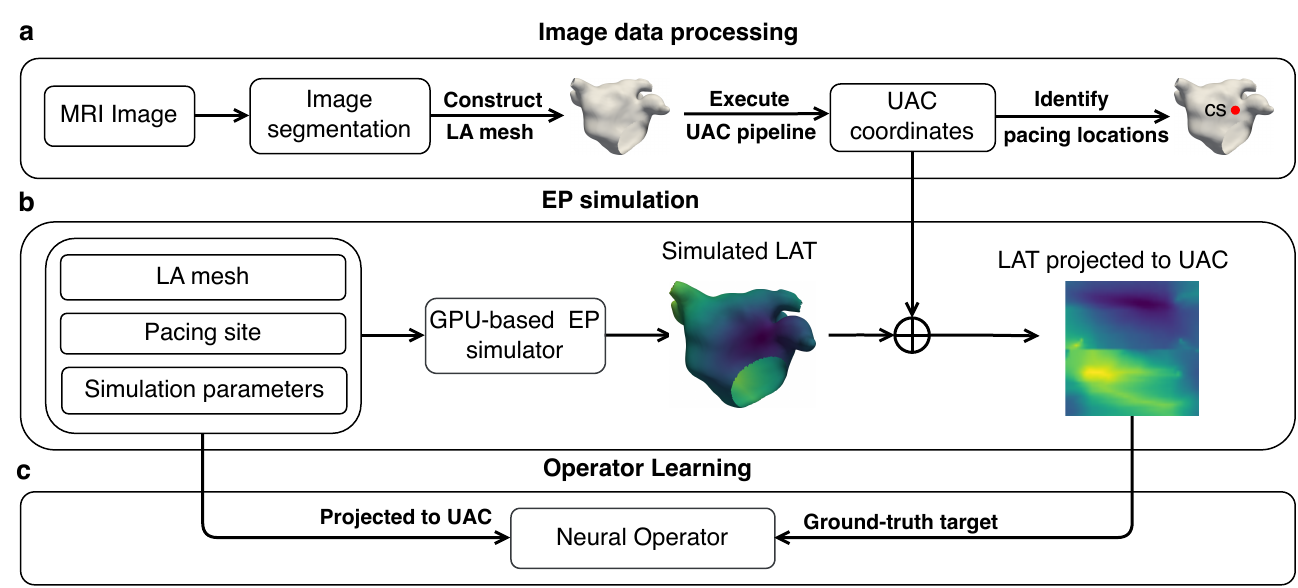}
  \caption{\textbf{Overview of the geometry-independent operator learning framework, shown for cardiac electrophysiology.}
  \textbf{a}, Preparation of left atrium (LA) mesh, mapping the 3D vertex coordinates defining the mesh to a 2D grid by Universal Atrial Coordinates (UAC), and identifying the pacing locations. 
 \textbf{b}, High-fidelity simulation on the original mesh and projection of output fields to the same coordinate grid to form training pairs. 
  \textbf{c}, Neural operator learning in the coordinate domain and reconstruction of predictions on the original mesh for evaluation and downstream analysis.}
  \label{fig:overview}
  
\end{figure}

In this study, cardiac electrophysiology serves as the primary use case, with the target field being the local activation time map (Fig.~\ref{fig:overview}). An additional kinematic deformation problem is used to demonstrate that the proposed framework can be applied unchanged to mechanically governed systems with different physics and outputs.

\subsection{Instantiation for cardiac electrophysiology}

\subsubsection{Geometries and cohorts}

The full dataset comprised 147 patient-specific left LA anatomies derived from two large-scale clinical cohorts. These anatomies served both as inputs to the EP simulations and as training and evaluation data for the neural operator models. To support both internal evaluation and cross-domain generalisation analysis, the 147 anatomies were partitioned into two distinct cohorts (Cohort A and Cohort B).

Cohort A, consisting of 100 AF LA samples (43 paroxysmal, 41 persistent, and 16 long-standing persistent), was obtained from the AF Recurrence Cohort \cite{roney2022predicting}, which was recruited in London, UK. LA geometries were segmented from late gadolinium-enhanced cardiac magnetic resonance (LGE-CMR) images acquired at end-diastole, yielding three-dimensional anatomical surfaces including the pulmonary veins and mitral annulus. Fibrotic tissue distribution was quantified using standard LGE intensity thresholding, providing patient-specific fibrosis maps that informed the EP simulations. This cohort served as the controlled domain for model development and internal validation.

Cohort B, containing 47 LA samples from AF patients recruited from Amsterdam, The Netherlands, was sourced from an independent clinical study \cite{lee2025regional}. The LA geometries in this cohort underwent the same segmentation and fibrosis quantification pipeline as those in Cohort A to maintain structural consistency.

As LGE-CMR does not resolve myofibre orientation, fibre direction vectors for anisotropic conduction were generated for all 147 anatomies using a model based on high-resolution ex-vivo Diffusion Tensor Magnetic Resonance Imaging (DTMRI) data \cite{pashakhanloo2016myofiber}. These fibre fields were subsequently incorporated into an atrial fibre atlas and projected onto each patient-specific geometry using the UAC system, as described by \cite{roney2021constructing}. 
 
Notably, Cohorts A and B were derived from two distinct and independent studies conducted by separate research groups operating within independent healthcare systems. This independence in data acquisition and analysis supports robust cross-domain evaluation. Moreover, despite adhering to equivalent segmentation and processing workflows \cite{solis2023evaluation, razeghi2020cemrgapp}, the two cohorts exhibit meaningful domain shifts. Variations arise from MRI acquisition parameters (scanner vendor, pulse sequences, and field strength), from different practices in anatomical trimming (particularly of pulmonary veins and the LA appendage), and from inter-operator differences in anatomical landmarking and boundary definitions. These protocol-level and anatomical choices shape the resulting geometries and fibrosis patterns, such that models trained exclusively on Cohort A typically encounter Cohort B as an out-of-distribution domain. These differences are not physiological changes. They are artefacts of data preparation, yet they change the geometric domain on which the operator is defined.

\subsubsection{Intrinsic coordinate projection and reconstruction}

The framework assumes the availability of an intrinsic coordinate mapping that provides a consistent parameterisation across heterogeneous geometries. Using such a mapping, spatial coordinates, fibre directions, and scalar fields defined on patient-specific meshes are projected onto structured grids of fixed resolution.
For the left atrium, UAC defines a bijective transformation
\[
\Phi_{\mathrm{LA}} : \mathcal{M} \rightarrow [0,1]^2,
\]
where \( \mathcal{M} \subset \mathbb{R}^3 \) denotes the atrial surface manifold. The mapping is constructed from anatomical landmarks, including the pulmonary vein junctions and mitral annulus, yielding a geodesic distance-based unfolding that aligns corresponding anatomical regions across patients \cite{uac}. An analogous intrinsic parameterisation is applied to ventricular geometries using the UVC system, which maps the three-dimensional ventricular volume to a structured coordinate domain.

Let \( f : \mathcal{M} \rightarrow \mathbb{R} \) denote a scalar field on the mesh (e.g., LAT), and let \( \boldsymbol{\xi}_i = \Phi(\mathbf{x}_i) \) be the corresponding UAC or UVC coordinates of mesh point \( \mathbf{x}_i \). A coordinate-domain representation \( \tilde{f} \) is constructed via spatial interpolation,
\[
\tilde{f}(\boldsymbol{\xi}) = \mathcal{I}\!\left(\{(\boldsymbol{\xi}_i, f_i)\}_{i=1}^{N}\right),
\]
where \( \mathcal{I}(\cdot) \) combines linear interpolation in densely sampled regions with nearest-neighbour interpolation near anatomical boundaries.

For atrial meshes, three-dimensional vertex coordinates \((x,y,z)\) are mapped to a grid in the UAC domain. Fibre orientations, originally defined per element, are converted to vertex-wise vectors using area-weighted averaging and normalised to unit length before projection. An equivalent procedure is applied in the UVC domain for ventricular geometries. After neural operator inference in the coordinate domain, predictions can be mapped back to the original patient-specific geometry. For each mesh point \( \mathbf{x} \in \mathcal{M} \) with coordinate \( \boldsymbol{\xi} = \Phi(\mathbf{x}) \), the predicted value is recovered by interpolating the learned field \( \hat{f}(\boldsymbol{\xi}) \).

\subsubsection{High-fidelity electrophysiology simulations}

To generate a large-scale, comprehensive dataset for training, we simulated LA electrical propagation using the widely adopted monodomain equation \cite{geselowitz1983bidomain} (more implementation details are provided in Supplementary Section~1, with an example of simulated wavefront propagation shown in Supplementary Fig.~1 and benchmark verification results in Supplementary Fig.~2). This biophysical model captures the electrical conduction through anisotropic cardiac tissue, balancing capacitive and ionic currents with diffusive current flow, governed by the longitudinal ($\sigma_l$) and transverse ($\sigma_t$) conductivities.

To ensure the generalisability of neural operators and to capture a wide range of conduction dynamics, the EP simulations were initiated using a distributed set of seven clinically relevant pacing sites in the LA: the Left Atrial Appendage (LAA), the Left Superior and Inferior Pulmonary Veins (LSPV, LIPV), the Right Superior and Inferior Pulmonary Veins (RSPV, RIPV), the Coronary Sinus (CS), and the Roof which is defined as the superior wall of the left atrium connecting the LSPV and RSPV \cite{romero2017atrial, moreira2008atrial, elmariah2013coronary}.

\begin{figure}[htbp]
    \centering
    \includegraphics[width=\linewidth]{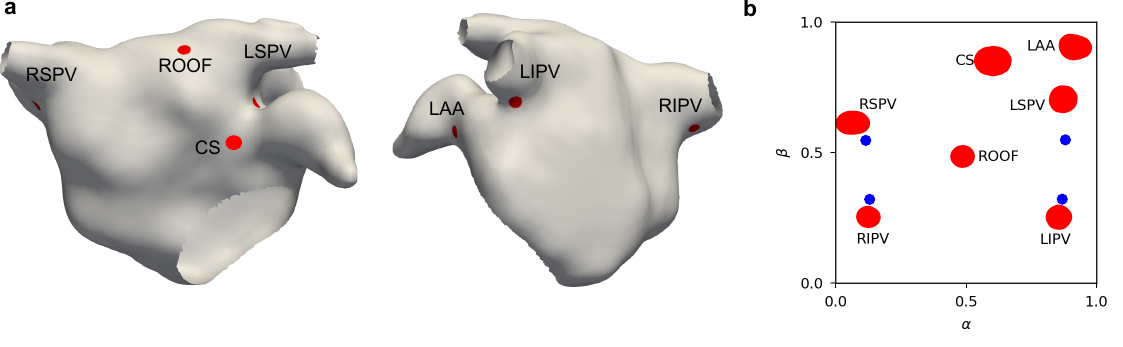}
    \caption{\textbf{Marking pacing sites on an LA.} \textbf{a}, Anterior and posterior views of an LA mesh where the seven pacing locations have been marked. \textbf{b}, Spatial distribution of these pacing sites in UAC coordinates. The blue regions indicate the anatomical openings (pulmonary veins and mitral valve). }
    \label{fig:uac_pacing}
\end{figure}

Pacing locations are defined across all patient anatomies using the UAC system. For each anatomical region, a 2D bounding box within the UAC domain was specified, delineating a subset of mesh vertices. The physical pacing centre is computed as the mean position of all vertices within that UAC-defined box. The stimulation region is subsequently defined by all mesh vertices located within a fixed Euclidean radius of $2.0\,\text{mm}$ from this central point, ensuring comparable stimulation volumes across all seven sites. The spatial distribution of these seven sites on a representative LA mesh is mapped onto the UAC coordinate system (Fig.~\ref{fig:uac_pacing}).

The primary and clinically significant output extracted from these simulations is the LAT map, which defines the time $t$ at which the transmembrane potential $V$ first crosses a predefined threshold ($V_{\text{th}}$, typically 0 mV) at any location $x$ in the cardiac domain:
\[
\mathrm{LAT}(x) = \min \left\{ t > 0 \ \middle| \ V(x, t) > V_{\text{th}} \right\},
\]

For each patient, we varied the longitudinal ($\sigma_l$) and transverse ($\sigma_t$) conductivities across seven clinically relevant pacing locations. At each pacing site, 300 combinations of $\sigma_l$ and the anisotropy ratio $\sigma_l/\sigma_t$ were sampled using Latin hypercube sampling \cite{mckay2000comparison}. The chosen range of $\sigma_l$ (0.1--0.4~S/m) is consistent with experimentally measured conduction velocities and previously reported myocardial conductivity values~\cite{fu2024progress}, while anisotropy ratios between 4 and 10 reflect the substantially faster fibre-aligned conduction observed in atrial tissue~\cite{kotadia2020anisotropic}. These values imply corresponding $\sigma_t$ in the range 0.01--0.1~S/m.

Each simulation was run for 600~ms to capture the full activation sequence, and the LAT map was recorded during runtime. Each simulation required approximately one minute on an NVIDIA A100 GPU, enabling the generation of 210{,}000 simulations for Cohort~A and 98{,}700 simulations for Cohort~B, yielding a combined total of 308{,}700 simulations. Executing these simulations across eight A100 GPUs reduced the total wall-clock time from an estimated 214 days (single GPU) to approximately 27 days. 

\subsubsection{Operator learning}
Within the proposed framework, we formulate LAT map prediction as a neural operator learning problem, in which the goal is to approximate a mapping from patient-specific anatomical and electrophysiological parameters to the corresponding spatial distribution of activation times. Formally, we aim to learn an operator
\[
\mathcal{G} : (A, \sigma, p) \mapsto \mathrm{LAT}(\mathbf{x}),
\]
where \(A\) denotes the atrial geometry, \(\sigma\) the conductivity tensor, \(p\) the pacing site location, and \(\mathrm{LAT}(\mathbf{x})\) the activation time at position \(\mathbf{x} \in \Omega\) where \(\Omega\) is the computational domain. This framework requires the generation of large and diverse training datasets that pair input configurations \((A, \sigma, p)\) with ground-truth LAT maps obtained from high-fidelity EP simulations.

\begin{figure}[ht]
  \centering
  \includegraphics[width=\textwidth]{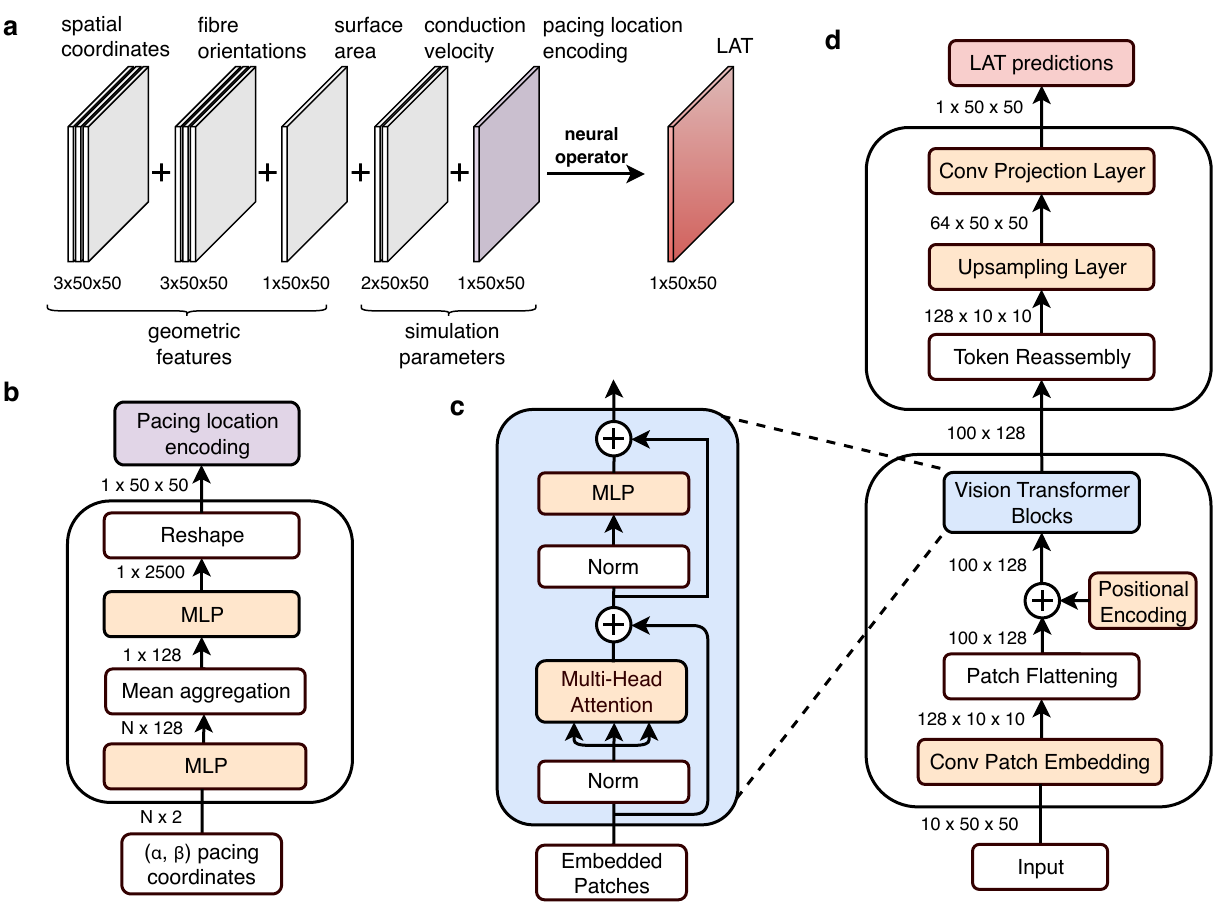}
  \caption{\textbf{Model architecture.}
  \textbf{a}, The construction of the input and output for neural operator learning, where patient-specific anatomical and physiological parameters are projected onto a structured grid via the Universal Atrial Coordinate system. The pacing location is converted from a point cloud representation into a spatial distribution on the same grid, enabling consistent integration with other input modalities. 
 \textbf{b}, A neural network module that maps a point cloud of shape $(N, 2)$, representing spatial input features such as pacing sites or anatomical landmarks, into a structured 2D grid of shape $(1, 50, 50)$. Here, $N$ refers to the number of vertices where stimulus (pacing) is applied. 
 \textbf{c}, The Vision Transformer block used in (d), which is the original ViT model proposed in \cite{vit}.
  \textbf{d}, Our full encoder-decoder architecture, which uses convolutional patch embedding and a Vision Transformer encoder to process the input grid, and then upsamples the latent features using a transposed convolution and projects them into the final LAT map using a convolutional output layer.}
  \label{fig:model}
  
\end{figure}

Our proposed architecture integrates a coordinate embedding module for point cloud projection with a Vision Transformer-based encoder and a convolutional decoder, enabling high-fidelity spatial field prediction from both sparse and structured physiological data. The model is designed to process heterogeneous inputs, including sparse point clouds (e.g., pacing coordinates) and spatially distributed scalar fields (e.g., tissue conductivity, surface area), by transforming them into a unified, fixed-resolution tensor representation suitable for transformer-based spatial modelling (Fig.~\ref{fig:model}).

\paragraph{Point-to-grid projection.}
To encode pacing (stimulation) locations provided as sparse, unordered two-dimensional point clouds of shape $(N, 2)$ where $N$ is the number of vertices in the pacing region, we introduce an embedding module that maps the point set into a 2D grid representation (Fig.~\ref{fig:model}b). This module follows the architectural design of PointNet \cite{qi2017pointnet}, applying a multi-layer perceptron (MLP) independently to each input point, followed by a permutation-invariant mean pooling operation to obtain a global descriptor of the stimulation pattern. This descriptor is then decoded via a learnable mapping to a 2D spatial grid of shape $(3, 50, 50)$. Unlike PointNet, which is optimised for classification and segmentation tasks, our adaptation is explicitly designed to produce spatially structured embeddings that are compatible with convolutional and transformer-based feature extractors. 

\paragraph{Multimodal feature integration.} 
The neural operator operates on a grid representation with multiple channels, which jointly encode mesh geometry and simulation parameters (Fig.~\ref{fig:model}a). These channels can be grouped into two categories. The first category encodes geometric and anatomical information. It consists of the UAC-projected three-dimensional vertex coordinates $(x, y, z)$ and the corresponding fibre direction vectors, yielding six spatial feature maps of size $(50, 50)$. In addition, the total surface area of the left atrium is included as a global anatomical descriptor and broadcast across the spatial domain to form a constant $(1, 50, 50)$ grid. The second category encodes simulation-specific parameters. This includes the longitudinal and transverse conduction velocities $(\sigma_l, \sigma_t)$, which are broadcast across the grid to form a $(2, 50, 50)$ feature map, as well as the pacing location, represented as a spatial distribution on a $(1, 50, 50)$ grid. The operator is trained to predict the corresponding LAT map, represented on a $(1, 50, 50)$ grid.

\paragraph{Spatial Representation Learning.}
To model spatial dependencies within the intrinsic coordinate domain, we employ a Vision Transformer (ViT) architecture \cite{vit} to encode the multi-channel input grid into contextualised latent representations. Attention-based models have demonstrated strong capability for learning solution operators of partial differential equations by capturing long-range interactions and global context \cite{li2023transformer,wu2024Transolver}. This capability is essential in our setting, where activation dynamics combine global wave propagation with sharp local transitions across the same spatial domain.

The input grid is first processed by a convolutional patch embedding layer \cite{wu2021cvt}, which partitions the domain into non-overlapping patches and projects each patch into a shared embedding space, yielding a sequence of latent tokens. Learnable positional embeddings are added to preserve spatial ordering within the tokenised representation. The token sequence is then processed by a stack of standard ViT blocks, each composed of multi-head self-attention and feed-forward sublayers, enabling the integration of global context and the modelling of long-range spatial interactions (Fig.~\ref{fig:model}c).

Following the encoder, the latent tokens are reshaped into a compact two-dimensional feature map. To recover full spatial resolution, we employ a lightweight decoder consisting of a transposed convolution layer for upsampling, followed by a convolutional projection head that maps latent features to the output field. The transposed convolution operation, originally introduced by \cite{zeiler2011adaptive} for adaptive deconvolutional networks and later popularised for convolutional network visualisation by \cite{zeiler2014visualizing}, enables learnable upsampling through the transpose of the convolution operator. This decoding stage ensures that the global representations learned by the transformer are translated back into a high-resolution spatial field suitable for pixel-wise prediction.

\paragraph{Sobolev Loss with Spatial Smoothness Regularisations}
To ensure our model produces physiologically consistent LAT maps, we employ an objective function grounded in the relative Sobolev $H^1$ norm. This choice enforces consistency not only in the predicted values but also in their spatial derivatives, a property essential for accurately modelling continuous and smooth conduction patterns.

To further refine the solution and mitigate non-physical, high-frequency artefacts while preserving sharp wavefront transitions, we incorporate two standard regularisation terms. The first is Total Variation (TV) regularisation $\mathcal{L}_{\mathrm{TV}}$, which penalises the $L^1$ norm of the image gradient. The second is a Laplacian regularisation $\mathcal{L}_{\mathrm{Laplacian}}$, which promotes harmonicity by penalising spatial curvature (full definitions are provided in Supplementary Section 4).

The overall objective function is defined as a weighted sum:
$$
\mathcal{L}_{\text{total}} = \mathcal{L}_{\mathrm{H}^1}(u, v) + \lambda \mathcal{L}_{\mathrm{TV}}(u) + \lambda\mathcal{L}_{\mathrm{Laplacian}}(u),
$$
where $\mathcal{L}_{\mathrm{H}^1}(u, v)$ is the primary loss between the predicted LAT map ($u$) and the ground truth ($v$), and the hyperparameter $\lambda$ controls the strength of the regularisation. This combination ensures that the model's predictions are both highly accurate and physiologically coherent.

\subsection{Instantiation for large deformation mechanics}

We consider kinematic deformation of non-linear elastic solids under load as a representative volumetric mechanics problem. The computational domain is defined as a truncated ellipsoidal shell, a canonical approximation historically used to model left ventricular geometry \cite{van1980application}. Although simplified, this geometry exhibits non-uniform curvature and provides a well-defined basal surface for the application of boundary conditions.

Inflation simulations were performed by fixing all degrees of freedom at the basal plane and applying a linearly increasing internal pressure to the inner surface. The resulting displacement field is governed by wall thickness, material stiffness, and applied pressure. The material behaviour was modelled using a hyperelastic transversely isotropic formulation based on the Guccione constitutive law \cite{guccione1991passive}. Fibre directions were assigned circumferentially, with incompressibility enforced via a bulk modulus of $\kappa = 1000,\mathrm{kPa}$. Fibre, cross-fibre, and fibre-sheet stiffness parameters were fixed at $b_f = b_t = b_{fs} = 18.48$.

A dataset of 1,000 geometries was generated by varying four parameters: inner cavity diameter, wall thickness, bulk stiffness, and maximum internal pressure. Parameter ranges are reported in Table~\ref{tab:inflation-param}. All simulations were performed using the Cardiac Arrhythmia Research Package (CARP) \cite{vigmond2003computational}, with each simulation executed on 32 CPU cores.

For operator learning, displacement magnitude fields were projected from the physical mesh onto a shared intrinsic coordinate system defined by universal ventricular coordinates \cite{bayer2018universal}. All fields were represented on a fixed $50 \times 50 \times 50$ grid in this coordinate domain and mapped back to the original geometry for evaluation.

\begin{table}[t]
\centering
\caption{Inflation simulation parameter ranges.}
\begin{tabular}{ll}
\hline
Parameter & Range \\
\hline
Inner cavity diameter (mm) & 38--58 \cite{lang2015recommendations} \\
Wall thickness (mm)        & 4--14 \cite{kawel2012normal} \\
Bulk stiffness (dimensionless) & 1--5 \\
Maximum internal pressure (kPa) & 1--10 \\
\hline
\end{tabular}

\label{tab:inflation-param}
\end{table}

\section*{Declarations}

\textbf{Funding} \\
This work is supported by the NIHR Imperial Biomedical Research Centre (BRC) and by the British Heart Foundation Centre of Research Excellence (RE/24/130023). SAN is supported by NIH R01-HL152256 and R01-HL162260, ERC PREDICT-HF453 (864055), BHF (RG/20/4/34803), EPSRC (EP/P01268X/1, EP/Z531297/1) and by the Technology Missions Fund under the EPSRC Grant EP/X03870X/1 \& The Alan Turing Institute. CR receives funding from the British Heart Foundation (RG/20/4/34803). We acknowledge computational resources and support provided by the Imperial College Research Computing Service. \\

\noindent
\textbf{Data availability} \\
Dataset~A used in this study is publicly available on Figshare under the permanent identifier \href{https://doi.org/10.6084/m9.figshare.30712559}{10.6084/m9.figshare.30712559}. Dataset~B contains sensitive information and cannot be publicly released due to ethical and privacy restrictions. Access to Dataset~B may be requested from the corresponding author and will be provided subject to institutional approval and a formal data-sharing agreement. \\

\noindent
\textbf{Code availability} \\
All code used to generate the results in this study is available at:
\href{https://github.com/sagebei/unified_framework_operator_learning}{https://github.com/sagebei/unified\_framework\_operator\_learning}. \\

\noindent
\textbf{Author contributions} \\
B.Z. developed the computational methodology, performed simulations, and drafted the manuscript. C.C. contributed to model development and data analysis. S.Q. assisted with numerical experiments and model development. M.B. contributed to software implementation and optimisation. A.W.C.L. prepared the data. C.R. contributed to manuscript editing and provided the simulation data. C.R., M.J.W.G. and L.H.G.A.H. provided experimental data. M.Q. assisted with analysing results. S.N. supervised the study, contributed to conceptualisation, and revised the manuscript. All authors reviewed the manuscript. \\

\noindent
\textbf{Competing interests} \\
The authors declare no competing interests.


\bibliography{sn-bibliography}

\end{document}